\documentclass{article}
\usepackage{graphicx}
\usepackage{hyperref}
\usepackage{amsfonts}

\title{Deep Learning and Mathematical Intuition: \\
A Review of (Davies et al. 2021)}
\author{
Ernest Davis \\
Dept. of Computer Science \\ New York University \\ New York, NY 10012 \\
{\small davise@cs.nyu.edu}}

\begin{document}
\maketitle
\begin{abstract}
A recent paper by Davies et al (2021) describes how deep learning (DL)
technology was used to find plausible hypotheses that have led to 
two original mathematical results: one in knot theory, one in
representation theory. I argue here that the significance and novelty of
this application of DL technology to mathematics is significantly
overstated in the paper under review and has been wildly overstated in
some of the accounts in the popular science press. In the knot theory
result, the role of DL was small, and a conventional statistical analysis
would probably have sufficed. In the representation theory result, the
role of DL is much larger; however, it is not very
different in kind from what has been done in experimental mathematics for
decades. Moreover, it is
not clear whether the distinctive features of DL that make it useful here
will apply across a wide range of mathematical problems.
Finally, I argue that the DL here
``guides human intuition'' is unhelpful and misleading; what the DL does
primarily does is to mark many possible conjectures as false and a few
others as possibly worthy of study.

Certainly the representation theory result represents an original and 
interesting application of DL to mathematical research, but its larger
significance is uncertain.
\end{abstract}

\section{Introduction}
A recent paper by Davies et al. (2021), published in {\em Nature}, discusses
two newly proved theorems --- one in knot theory, one in representation 
theory\footnote{For AI readers: ``representation theory'' here is a subfield
of abstract algebra. It has essentially no connection to either the knowledge
representation used in symbolic AI, or the representations studied in 
machine learning.} --- where the authors were to some extent guided by
computational experiments carried out using deep learning.\footnote{The
proofs of the theorems are published separately, in (Davies, Lackenby,
Juhasz, and Tomasev, 2021) and (Blundell et al., 2021), and are included
in the supplemental material.}

The paper has gathered a great deal of attention in the popular scientific
press, in uncritical articles with overheated headlines, such as 
``DeepMind AI collaborates with humans on two mathematical breakthroughs'' in 
{\em NewScientist} (Sparkes, 2021); 
``Maths researchers hail breakthrough 
in applications of artificial intelligence'' in  {\em phys.org,} 
(U.~Sydney, 2021); ``Researchers create AI that can create brand new math
theorems'' in {\em iflscience.com} (Spalding, 2021).
An article
in {\em ScienceAlert\/} (Neild, 2021) claims 
``AI is discovering
patterns in pure mathematics that have never been seen before \ldots 
We can add suggesting and proving mathematical theorems to the long list 
of what artificial intelligence is capable of.''  An article in LiveScience
(Pappas, 2021) describes the results as ``the first-ever important advances 
in pure mathematics generated by artificial intelligence.'' 

On careful examination, however, this excitement is excessive and the
articles are misleading.  The AI in these two mathematical research project
neither suggested nor proved any theorems. 
The use of AI in these 
projects is not very different in kind from what has been done in experimental
mathematics (Borwein and Bailey, 2008) for decades. 
In the case of the knot
theory result, though not the representation theory result, the
insight provided by deep learning could have been achieved through 
more traditional statistical analysis. The claim that the AI was used to
``guide human intuition'' is misleading and unhelpful; a more accurate claim
is that the experimentation indicated that some specific conjectures might be
true and therefore deserve study while others almost certainly were not, and
therefore should be abandoned.

Before going further, let me make two disclaimers. The first has to do
with my own limitations. The work here draws on three different fields, and
I am not an expert in any of them. I have some knowledge of machine learning
generally and deep learning specifically, but I am not an expert. I know
very little about knot theory and almost nothing at all about representation
theory. So it is certainly possible that my account below has some misstatements
or misjudgments. If you find a mistake, please do let me know.

Second, let me clearly set out the limits of what I am claiming here. I am
not at all arguing against the value of computational experiments in 
mathematical research or against the development of AI tools for
mathematics; there is a lot of excellent, exciting work going on in both
directions. I am not arguing against the significance of these new mathematical 
results as mathematics; I am in no position to judge that. I am not arguing
that the {\em Nature\/} paper misrepresents how the AI tools were used or
how helpful the mathematicians involved found them; I am working from the
assumption that that is all accurately set forth.

What I want to do in this review is to lay out clearly the role of AI in
these projects and to discuss the implications of that role. Since
many lay readers and even many people involved in machine learning technology
have little idea of research in pure mathematics, and since many mathematicians
do not follow machine learning closely, claims such as the ones I have quoted
above can easily be accepted uncritically, leading to miscommunication and
misunderstanding.

Ordinarily, when someone praises a tool that they have been using ---
a gardener praises a particular hedge trimmer, a mathematician says that 
they constantly use Brouwer's fixed point theorem --- 
there is no reason to be at all skeptical. In this case, however, there
is a perverse incentive to focus the role of the DL in this research,
not just for the ML specialists from Deep Mind, but  even for the 
mathematicians. The remarkable fame that these works have achieved ---
the prestigious article in {\em Nature\/}, the excitement in all kinds
of science venues across the internet --- has nothing to do with their 
merits  as pure mathematics, and everything to do with the fact that AI
was involved. Everyone involved, therefore, has an interest in highlighting
and even exaggerating the extent and impact of that involvement.

\section{The knot theory result}
My discussion of the mathematical aspects of the two results will be minimal;
only as much as is necessary to explain the role of the AI in achieving them.
For a little more detail, go to (Davies et al., 2021); for a lot more detail,
look up the works cited there.

A {\em knot\/} in mathematical parlance is a closed loop in a
three-dimensional space.  There are many different numerical
features of a knot
that can be defined that describe different geometric characteristics. 
These features are categorized into
a number of different categories, corresponding to the kind of math used
to define them. Two of the most important categories are {\em algebraic
invariants} and {\em hyperbolic invariants}.
The goal of the research here was to find a relation between one particular
algebraic invariant, called the {\em signature\/} 
and a number of hyperbolic invariants; that would exciting because
it would be a connection between two completely separate approaches to
characterizing a knot.

The research project proceeded along a number of steps:

{\bf Step 1:} The mathematicians selected about 12 hyperbolic invariants as
promising candidates.\footnote{The description in (Davies et al. 2021) is
not entirely clear on this point. The list of invariants given in the 
section Methods/Topology/Data-Generation is slightly different from the list
in figure 3.}

{\bf Step 2:} They created a data set with 2.7 million examples of knots, 
labelled with the signature and these 12 hyperbolic invariants.

{\bf Step 3:} Using the data set, they trained a 
neural network (3 hidden layers, 300 units in each, fully connected)
with the task of predicting the signature from the hyperbolic invariants.

{\bf Step 4:} They observed that the trained NN could predict the signature from
the inputs with an accuracy of 78\%. This indicated that, indeed, there is
a relation.

{\bf Step 5:} Using standard techniques for analyzing neural networks, 
they determined that they can reduce the input to three hyperbolic invariants 
--- the real and imaginary parts of the meridional translation and the 
longitudinal translation --- without losing much accuracy.

{\bf Step 6:} Thinking about these particular quantities, the mathematicians
decided that it would be reasonable to combine them into a new, geometrically
meaningful, quantity, the `natural slope'. 
If $\mu$ is the meridional translation 
(a complex number, thus a pair of real numbers) and $\lambda$ is the
longitudinal translation, then slope = Re($\lambda/\mu$). 

{\bf Step 7:} They plotted the signature against the slope. It turned
out that there was an approximate linear relation: the signature is 
approximately half the slope. Using the slope as the sole predictive feature
for the signature gave an accuracy of 78\%; i.e. the slope constituted
all the predictive power in the twelve hyperbolic features combined.

{\bf Step 8:} They formulated an initial hypothesis:
\[ | 2\sigma(K) - \mbox{slope}(K)| < c_{1}\mbox{vol}(K) + c_{2} \]
where $K$ ranges over knots, $\sigma$ is the signature, vol($K$)
is the volume (a fourth hyperbolic parameter), and $c_{1}$ and $c_{2}$
are constants.

{\bf Step 9:} They realized that there might be counter-examples to the
initial hypothesis over a particular category of knots. They generated a data
set with about 36,000 knots in this category. They indeed found some 
counter-examples.

{\bf Step 10:} They formulated a revised hypothesis 
\[ | 2\sigma(K) - \mbox{slope}(K)| < c\, \mbox{vol}(K)/\mbox{inj}(K)^{3} \]
where inj($K$) is the injectivity radius, a fifth hyperbolic parameter,
which can be a small real number.

{\bf Step 11:} They were able to prove the revised hypothesis.

Note, first, that machine learning was used only in steps 4 and 5, and its
sole outcome is to determine that three particular hyperbolic 
parameters are largely predictive of the signature. Step 7 involved doing a
linear regression. Steps 2 and 9, the creation of two data sets, involved highly
specialized non-AI software, almost certainly more sophisticated computationally
and mathematically than the neural network, quite possibly more demanding
computationally (the computation of some of these parameters can be 
very difficult.) The other steps were all carried out by the human 
mathematicians.

Second, at the end of the day, all that the deep learning component 
accomplished was to
isolate a mathematically simple relationship between four real-valued numerical
parameters. If we write the complex meridional translation $\mu$ in the form
$\mu = a+bi$, then the formula for slope becomes slope = Re($\lambda/a+bi$)
= $\lambda a /(a^{2}+b^{2})$. The relation detected by the neural network 
therefore has the form $\sigma \approx \lambda a / 2 (a^{2}+b^{2})$ ---
linear in $\lambda$ and a simple function of $a$ and
$b$. Again, the DL did not even extract the form of the relation; all it
did was to report that $\sigma$ could be predicted with 78\% accuracy from
$\lambda$, $a$, and $b$.

Broadly speaking, this knot theory problem is not actually the kind of 
problem where DL typically outshines other machine-learning or statistical
methods. DL's strength is in cases like vision or text where each instance 
(image or text) has a large numbers of low-level input features, it is hard to 
reliably identify high-level features, and the function relating the input
features to the answer is, as far as anyone can tell, immensely complex, with
no small subset of the input features being at all determinative.
With only twelve input features, of which only three are relevant, with
a simple mathematical approximate relation involved, and with three 
million data points, it is hard to see why a neural network with 200,000 
parameters would be
the method of choice; simple, conventional statistical methods or a support
vector machine would be more suitable.

\section{The representation theory result}
Like the knot theory result, the representation theory result involves 
connecting two different views of a mathematical object. The objects
involved in representation theory are much more abstract than those in
knot theory, and I will not make the slightest attempt to explain them; 
I will merely quote them.  (The account of the math in (Davies et al. 2021)
is likewise noticeably sketchier for the representation theory than the 
knot theory, for the same reason.)

Symmetric groups are a very important category of mathematical structure; 
generally speaking, they describe the ways that a set of some kind can
be mapped into itself. If you have two elements of a symmetry groups,
then the relation between those can be characterized in two different ways:
the unlabelled Bruhat interval graph and the Kazhdan-Lusztig (KL) polynomial.
The Bruhat interval graph 
rapidly becomes quite complex, even for fairly simple elements; the KL 
polynomial is much simpler. For instance, in the examples they used,
some of the graphs had 9! = 362,880 nodes, whereas the polynomials were at
most of degree four.


A long-standing conjecture, known as the ``combinatorial invariance'' 
conjecture states that the KL polynomial can be computed from the 
Bruhat interval graph.  One of the obstacles to proving this is that the 
graphs are so large that it is hard for researchers to get an intuitive 
feeling for their structure.

What the project here accomplished was to prove a result
that is a major step toward the full conjecture: There is a way to
pre-process the interval graph, called ``a hypercube decomposition along
its extremal reflection'', and once you have done that, then the KL-polynomial
can be computed using a simple formula. The only problem is, there can be
many different hypercube decompositions along extremal reflections, and all 
that Blundell et al. have proved is that {\em one\/} of these will work;
they haven't given a method of finding the right one. They have now formulated
the further conjecture that, in fact, {\em any\/} hypercube decomposition
along extremal reflections will work, and their experiments bear that out,
but they have not proved it. If they can prove that, then that will settle the
combinatorial invariance conjecture.

The project involved the following steps:

{\bf Step 1:} They designed a specialized, comparatively shallow (4 propagation
steps), deep learning
system that takes as input an encoding of the interval graph and produces
as output the coefficients of the KL polynomial. The architecture of the
network was designed to be ``algorithmically aligned'' with existing knowledge
about techniques for computing KL-polynomials from ``labelled Bruhat 
intervals'' i.e. Bruhat intervals with some extra information.

{\bf Step 2:} They created a training set of 24,000 examples, labelled by
the Bruhat interval graph and the KL polynomial.

{\bf Step 3:} They trained the network on the training set, attaining accuracies
for the four coefficients ranging from 63\% to 98\%. This success gave some
encouragement that the Bruhat interval graph does indeed determine the 
polynomials.

{\bf Step 4:} ``By experimenting on the way in which we input the Bruhat 
interval to the network, it became apparent that some choices of graphs 
and features were particularly conducive to accurate predictions. In 
particular, we found that a subgraph inspired by prior work may be 
sufficient to calculate the KL polynomial, and this was supported by 
a much more accurate estimated function.'' Accuracies on this subgraph ranged
between 95.6\% and 99\% for the different coefficients.

{\bf Step 5:} By analyzing of the states of the DL network generated in
the experiments of step 4, they determined that edges in the Bruhat graph
corresponding to ``extremal reflections'' tended to be particularly important
in the successful experiments. The authors state that they did not at all
anticipate this.

{\bf Step 6:} They used the information gathered in steps 4 and 5 to formulate
a conjecture that they were able to prove.

Clearly the role of deep learning is much greater here than in the knot theory
work. In particular my comment about the knot theory work, that other kinds of
statistical analysis would have worked as well, does not at all apply here.
The number of input features is enormous, and the relation between the
input feature and the output is complex.

However, a number of points should be observed.

In step 3, the DL provided little more than general encouragement that the
project on the whole was on the right track and that the deep learning system
might be able to detect the relation of the KL polynomial to the 
In step 4, it distinguished successful from
unsuccessful experiments with different forms of input. Presumably all
these experiments were carefully selected by the mathematicians as possibly
promising avenues of research.  It would be interesting to know how many
such experiments were run and how many were successful; i.e. how much useful
pruning did the DL provide here?  In step 5, analysis of the DL identified 
certain edges as particularly important; here the DL did, indeed, provide
information that the mathematicians had not anticipated.

Unlike the knot theory project, which used a generic DL architecture,
the neural network was carefully designed to fit deep mathematical 
knowledge about the problem. Moreover, the DL worked {\em much\/} better,
with something like 1/40th the error rate, on pre-processed data than on
the original data. 
This cuts both ways. On the one hand, in the 
past, critics of DL, including myself, have often objected that it is difficult
to incorporate domain knowledge; this cuts against that criticism. On the
other hand, enthusiasts for DL have often praised DL as a ``plug-and-play''
learning methodology that can be thrown at raw data for 
whatever problem comes to hand; this cuts against that praise.

Deep mathematical
knowledge of the problem was required here at practically every stage: In 
designing the DL architecture in step 1, in creating the data set in step 2,
in choosing the experiments in step 4, in interpreting them in step 5. and
of course in proving the theorem in step 6. 

In both projects, the DL was entirely in the dark about the larger mathematical
setting of the problem it was solving. In the knot theory project, it was not
given the actual geometric structure of the knots, just the invariants as a
collection of uninterpreted numbers. In the representation theory project, it
was not given the group or the group elements, just the Bruhat graph and the
KL polynomials.

In applications of deep learning of this kind, success may depend critically
on the way in which the training data is generated and the way that the
mathematical structures are encoded, in ways that are quite specific
to a particular task (Charton, 2021). Finding the best way to generate
and encode data involves a mixture of theory, experience, art, and
experimentation. The burden of all this lies on the human expert. Deep
learning can be a powerful tool, but it is not always a robust one.

In view of all this, it seems to me that DL as used here is best viewed as 
another analytic tool in the toolbox of experimental mathematics rather than 
as a fundamentally new approach to mathematics. How powerful a tool it 
is and how broadly
applicable remains to be seen. Experimental mathematics in general can only be
applied to certain kinds of mathematical questions, and this technique is 
further limited.  Davies et al. (2001) themselves express hesitation
on that:  

\begin{quote}
There are limitations to where this framework will be useful  --- 
it requires the ability to generate large datasets of the representations 
of objects and for the patterns to be detectable in examples that are 
calculable. Further, in some domains the functions of interest may be 
difficult to learn in this paradigm. 
\end{quote}

Additionally, there are examples where other analysis tools are more effective.

So far, we have one quite impressive example of the use of DL in mathematical
research and one that is much less impressive. The history of AI has had its 
share of one-offs; remarkable successes that were achieved on one particular
task that seemed promising but could never be matched on a second task. 

\section{``Guiding Intuition''}
Finally, ``guiding intuition'' seems to me an seriously
inaccurate description of the 
assistance that mathematicians have gained, or can hope to gain, from this
use of DL systems.
The word ``intuitive'' is a useful, though informal, one
in mathematical writing.\footnote{Of course, the phrase ``intuitively obvious'' 
is often abused, either to
bully students into quiet acceptance or as a euphemism for ``I haven't thought it
through but I'm pretty sure it's right.''}  Because it is a useful word, it is
important not to misuse it. In the mathematical setting, the word 
``intuitive'' means that a concept or a proof can be
grounded in a person's deep-seated sense of familiar domains such as numerosity, 
space, time, or motion, or in some other way ``makes sense'' or ``seems right''
in a way that does not involve explicit calculation or step-by-step reasoning.
Math teachers work hard
to try to give their students an intuitive understanding of concepts and proofs
by tying them to the familiar; this can involve diagrams, videos, examples
from everyday life, and similar tools. Mathematicians and math students work hard
to achieve an intuitive understanding by similar means. Contrary to the dictionary
definition, ``consisting in immediate apprehension, without the intervention 
of any reasoning process'', an intuitive grasp of a mathematical concept often
requires hard mental work to attain. Often, gaining an intuitive grasp requires
working through multiple specific examples.
But no teacher ever tries to guide their students' intuitions by telling them,
``I have looked at millions of examples, and I can report that there is a 
pattern that works X\% of the time, mostly on the basis of features A,B,C'' 
which is what the DL tells the mathematician,
and all that the DL tells the mathematician. The mathematicians in these
projects may well have attained an intuitive understanding of the concepts they
have defined, the theorems they have prove, and the conjectures they have put 
forward, but they did not get that from the DL. What the DL did was to give
them some advice as to which features of the problem seemed to be important
and which seemed unimportant. That is not to be sneezed at, but it should not
be exaggerated.

\subsection*{Acknowledgements}
Thanks to Fran\c{c}ois Charton for helpful suggestions.

\subsection*{References}
Blundell et al. (2021) ``Towards combinatorial invariance 
for Kazhdan-Lusztig polynomials.''  Arxiv preprint 2111.15161
\url{https://arxiv.org/abs/2111.15161}. 

\noindent
Borwein, J. and Bailey, D. (2008). {\em Mathematics by Experiment.}
CRC Press.
  
\noindent
Charton, F. (2021). ``Linear algebra with transformers.'' ArXiv preprint 
2112.01898 \\
\url{https://arxiv.org/abs/2112.01898}

\noindent
Davies et al. (2021). ``Advancing mathematics by guiding human intuition with
AI.'' {\em Nature}, {\bf 600}: 70-74. \\
\url{https://www.nature.com/articles/s41586-021-04086-x}

\noindent
Davies, A., Lackenby, M., Juhasz, A. and Tomašev, N. ``The signature and cusp 
geometry of hyperbolic knots.'' Arxiv preprint 2111.15323. \\
\url{https://arxiv.org/abs/2111.15323}

\noindent
Neild, D. (2021). ``AI is discovering
patterns in pure mathematics that have never been seen before.''
{\em ScienceAlert,} Dec. 4, 2021. \\
\url{https://www.sciencealert.com/ai-is-discovering-patterns-in-pure-mathematics-that-have-never-been-seen-before}

\noindent
Pappas, S. (2021). ``DeepMind cracks 'knot' conjecture that 
bedeviled mathematicians for decades.'' {\em livescience.com} \\
\url{https://www.livescience.com/deepmind-artificial-intelligence-pure-math}

\noindent
Spalding, K. (2021). 
``Researchers create AI that can create brand new math
theorems'' {\em iflscience.com}. \\
\url{https://www.iflscience.com/editors-blog/researchers-create-ai-that-can-invent-brand-new-math-theorems/}

\noindent
Sparkes, M. (2021) 
``DeepMind AI collaborates with humans on two mathematical breakthroughs.'' 
{\em NewScientist}, Dec. 1, 2021. \\
\url{https://www.newscientist.com/article/2299564-deepmind-ai-collaborates-with-humans-on-two-mathematical-breakthroughs/}

\noindent
University of Sydney (2021). ``Maths researchers hail breakthrough 
in applications of artificial intelligence.'' {\em phys.org,} 
Dec. 1, 2021. \\
\url{https://phys.org/news/2021-12-maths-hail-breakthrough-applications-artificial.html}

\end{document}